\documentclass[sigconf]{acmart}

\usepackage{booktabs} 
\usepackage{amsmath}
\usepackage[linesnumbered, ruled]{algorithm2e}
\usepackage{subcaption}
\usepackage[font=small,labelfont=bf,tableposition=top]{caption}
\edef\oldtt{\ttdefault}
\usepackage[scaled]{beramono}
\usepackage{tabularx}
\usepackage[T1]{fontenc}
\renewcommand*\ttdefault{\oldtt}
\newcommand{\bera}[1]{{\small\fontfamily{fvm}\selectfont #1}}
\usepackage{enumitem}
\usepackage{cleveref}
\usepackage{setspace}
\usepackage{afterpage}
\usepackage{float}
\theoremstyle{definition}

\SetCommentSty{mycommfont}
\makeatletter
\newcommand\footnoteref[1]{\protected@xdef\@thefnmark{\ref{#1}}\@footnotemark}
\makeatother
\setcopyright{rightsretained}

\acmDOI{10.475/123_4}

\acmISBN{123-4567-24-567/08/06}

\acmConference[Submitted to ECML/PKDD 2018]{}{}{}
\acmYear{2018}
\copyrightyear{2018}

\acmArticle{4}
\acmPrice{15.00}

\begin{document}
	\title{MetaBags: Bagged Meta-Decision Trees for Regression}

	\author{Jihed Khiari}
	\affiliation{%
		\institution{NEC Laboratories Europe}
	}
	\email{jihed.khiari@neclab.eu}
	
	\author{Luis Moreira-Matias}
	\affiliation{%
		\institution{NEC Laboratories Europe}
	}
	\email{luis.moreira.matias@gmail.com}

	\author{Ammar Shaker}
	\affiliation{%
		\institution{NEC Laboratories Europe}
	}
	\email{ammar.shaker@neclab.eu}	
	
	\author{Bernard \v{Z}enko}
	\affiliation{%
		\institution{Josef Stefan Institute}
	}
	\email{bernard.zenko@ijs.si}
	
	\author{Sa\v{s}o D\v{z}eroski}
	\affiliation{%
		\institution{NEC Laboratories Europe}
	}
	\email{saso.dzeroski@ijs.si}
	
	\renewcommand{\shortauthors}{Khiari et al.}
	
	\begin{abstract}
	Ensembles are popular methods for solving practical supervised learning problems. They reduce the risk of having underperforming models in production-grade software. Although critical, methods for learning heterogeneous regression ensembles have not been proposed at large scale, whereas in classical ML literature, stacking, cascading and voting are mostly restricted to classification problems. Regression poses distinct learning challenges that may result in poor performance, even when using well established homogeneous ensemble schemas such as bagging or boosting.
	
	In this paper, we introduce \texttt{MetaBags}, a novel, practically useful stacking framework for regression. \texttt{MetaBags} is a meta-learning algorithm that learns a set of meta-decision trees designed to select one base model (i.e. \emph{expert}) for each query, and focuses on inductive bias reduction. A set of meta-decision trees are learned using different types of meta-features, specially created for this purpose. Each meta-decision tree is learned on a different data bootstrap sample, and, given a new example, selects a suitable base model that computes a prediction. Finally, these predictions are aggregated into a single prediction. This procedure is designed to learn a model with a fair bias-variance trade-off, and its improvement over base model performance is correlated with the prediction diversity of different experts on specific input space subregions. The proposed method and meta-features are designed in such a way that they enable good predictive performance even in subregions of space which are not adequately represented in the available training data.
	
	An exhaustive empirical testing of the method was performed, evaluating both generalization error and scalability of the approach on synthetic, open and real-world application datasets. The obtained results show that our method significantly outperforms existing state-of-the-art approaches.
	\end{abstract}
    
	
	\maketitle

	\section{Introduction}
\emph{Ensemble} refers to a collection of several models (i.e., experts) that are combined to address a given task (e.g. obtain a lower generalization error for supervised learning problems) \cite{mendes2012}. Ensemble learning can be divided in three different stages \cite{mendes2012}: (i) base model \emph{generation}, where $z$ multiple possible hypotheses $\hat{f}_i(x), i \in \{1..z\}$ to model a given phenomenon $f(x)=p(y|x)$ are generated; (ii) model \emph{pruning}, where $c \le z$ of those are kept and the others discarded; and (iii) model \emph{integration}, where these hypotheses are combined to form the final one, i.e. $\hat{F}\big(\hat{f}_1(x),...,\hat{f}_c(x)\big)$. Naturally, the whole process may require a large pool of computational resources for (i) and/or large and representative training sets to avoid overfitting, since $\hat{F}$ is also estimated/learned on the (partial or full) training set, which was already been used to train the base models $\hat{f}_i(x)$ in (i). Since the pioneering Netflix competition in 2007 \cite{bell2007} and the coincident introduction of cloud-based solutions for data storing and/or large-scale computing purposes, ensembles have been increasingly often used for industrial applications. A good illustration of such a trend is \emph{Kaggle}, the popular competition website, where, during the last five years, 50+\% of the winning solutions involved at least one ensemble of multiple models \cite{kaggle2018a}. 

Ensemble learning builds on the principles of committees, where there is typically never a single expert that outperforms all the others on each and every query. Instead, we may obtain a better overall performance by \emph{combining} answers of multiple experts \cite{schaffer1994}. Despite the importance of the combining function $\hat{F}$ for the success of the ensemble, most of the recent research on ensemble learning is either focused on (i) model generation and/or (ii) pruning \cite{mendes2012}.

\begin{figure*}[!t]
      \centering
	\begin{subfigure}[b]{0.32\linewidth}
		\centering
		\includegraphics[trim=0.5cm 0.3cm 0.25cm 0.1cm, width=\linewidth,clip=TRUE]{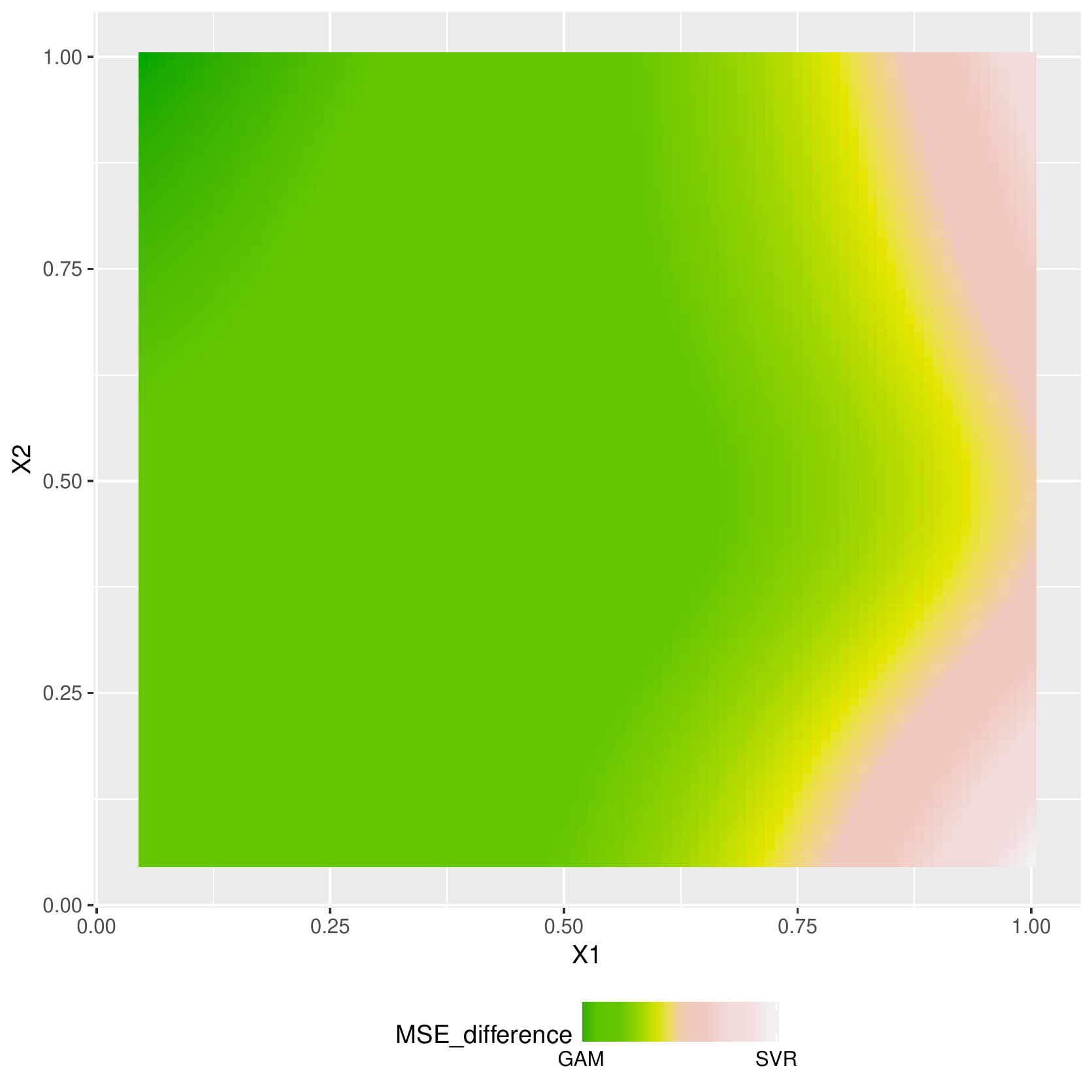}
		\label{fig1-1} 
		\vspace{-0.45cm}\caption{\texttt{GAM} vs. \texttt{SVR}.}
	\end{subfigure}\hspace{0.03cm}
	\begin{subfigure}[b]{0.315\linewidth}
		\centering
		\includegraphics[trim=0.8cm 0.85cm 0.7cm 0.8cm, width=\linewidth,clip=TRUE]{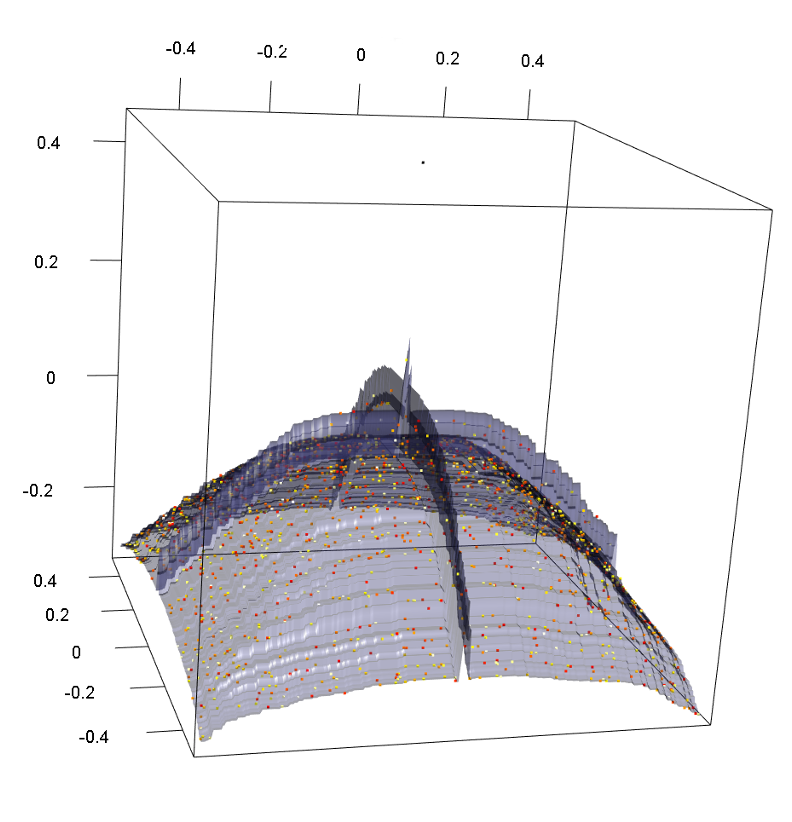}
		\label{fig1-2} 
		\vspace{-0.5cm}\caption{\texttt{GB} w. Target outlier.}
	\end{subfigure} 
	\begin{subfigure}[b]{0.315\linewidth}
		\centering
		\includegraphics[trim=1.8cm 0.85cm 1.0cm 2.1cm, width=\linewidth,clip=TRUE]{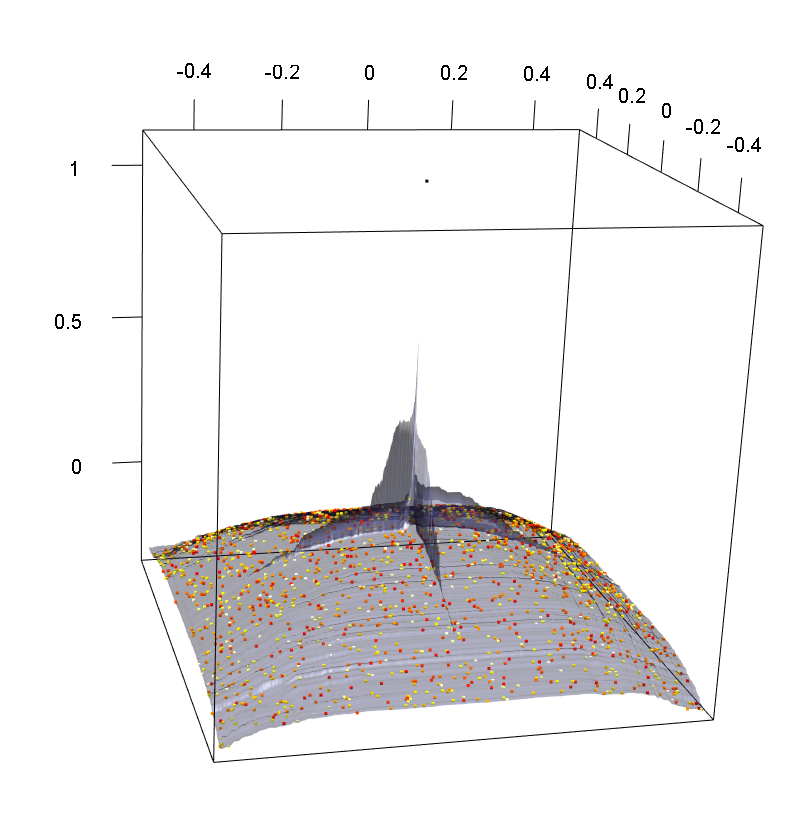}
		\label{fig1-3} 
		\vspace{0.1cm}\vspace{-0.6cm}\caption{\texttt{RF} w. Target outlier.}
	\end{subfigure}\newline
    \begin{subfigure}[b]{0.365\linewidth}
		\centering
		\includegraphics[trim=0.8cm 0.85cm 0.7cm 0.1cm, width=\linewidth,clip=TRUE]{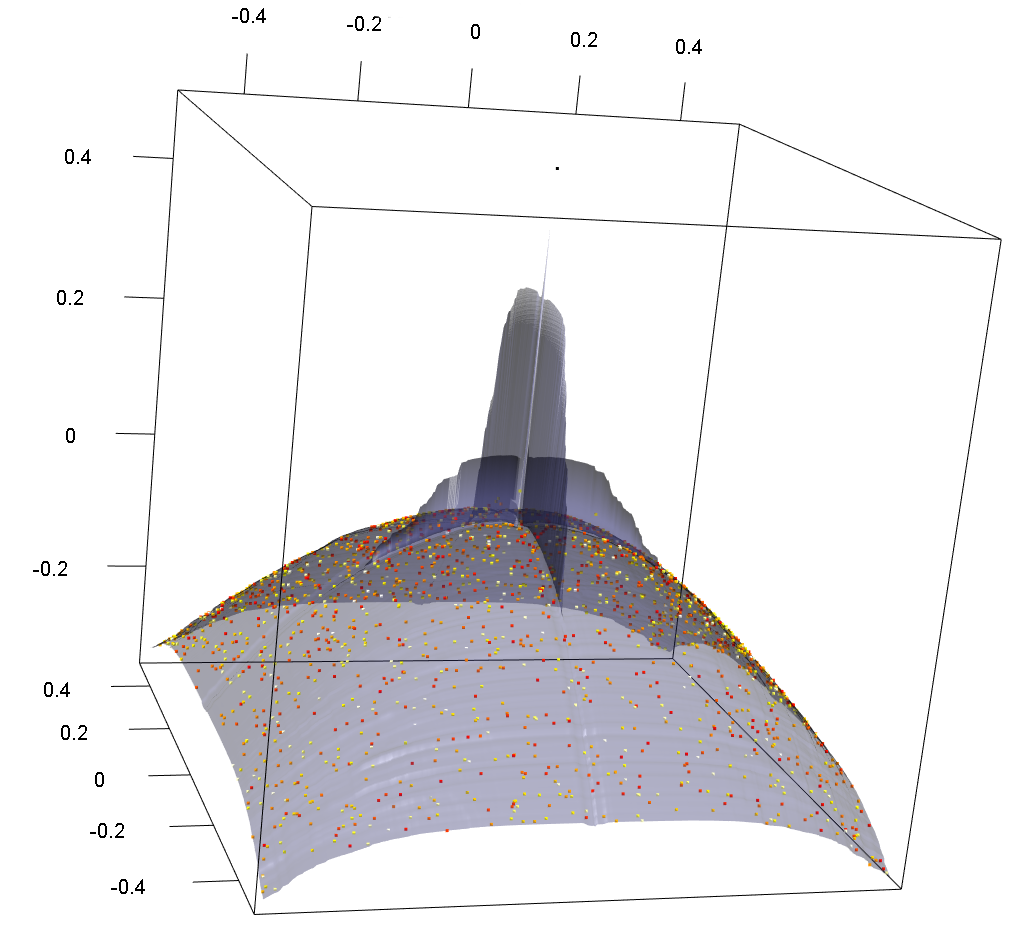}
		\label{fig1-4} 
		\vspace{-0.5cm}\caption{\texttt{LS} w. Target outlier.}
	\end{subfigure} 
	\begin{subfigure}[b]{0.29\linewidth}
		\centering
		\includegraphics[trim=1.8cm 0.85cm 1.0cm 0.1cm, width=\linewidth,clip=TRUE]{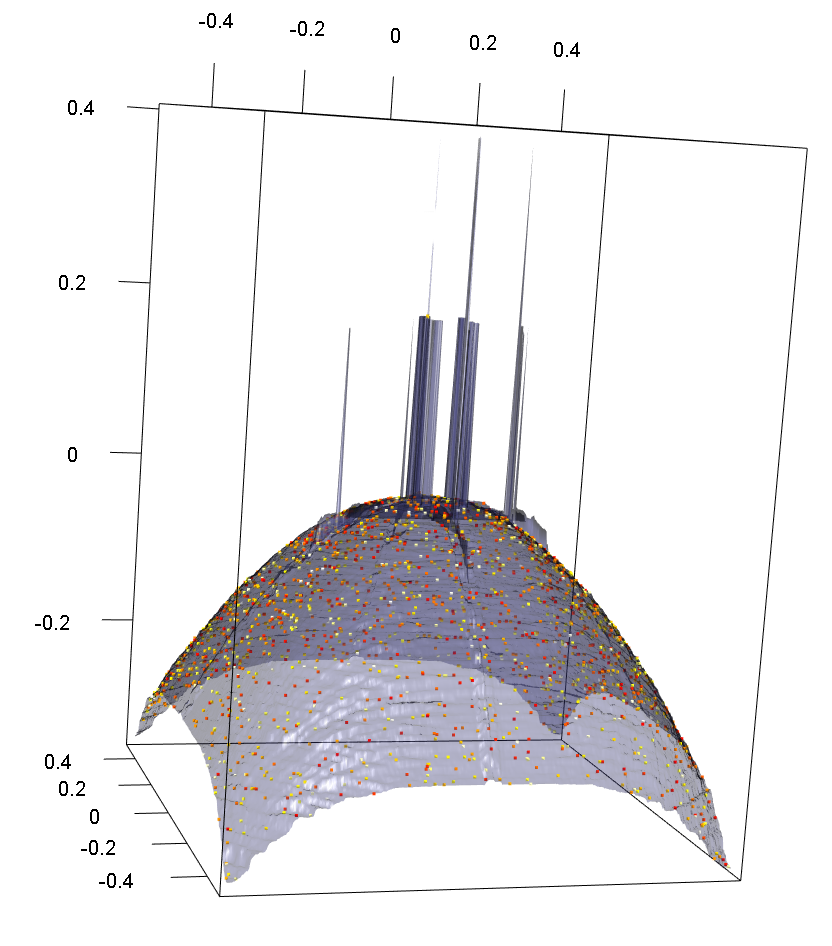}
		\label{fig1-5} 
		\vspace{0.1cm}\vspace{-0.6cm}\caption{\texttt{DS} w. Target outlier.}
	\end{subfigure}
    	\begin{subfigure}[b]{0.32\linewidth}
		\centering
		\includegraphics[trim=0.2cm 0.65cm 0.6cm 0.5cm, width=\linewidth,clip=TRUE]{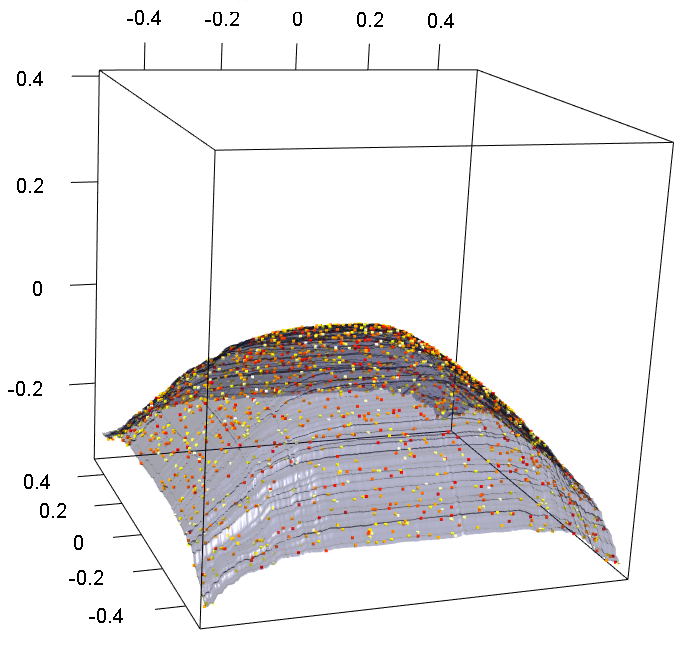}
		\label{fig1-6} 
		\vspace{0.1cm}\vspace{-0.6cm}\caption{Original surface.}
	\end{subfigure}
	\caption{Illustration of distinctive issues in regression problems on a syntethic example. In all experiments, we generate 1k training examples for the function $y=(x_1^4+x_2^4)^{\frac{1}{2}}$. In (a), $x_1,x_2$ training values are sampled from a uniform distribution constrained to $\in [0,0.8]$, while the testing ones are $\in [0,1]$. Panel (a) depicts the difference between RMSE between the two tested methods, \texttt{GAM} and \texttt{SVR}, where the hyperparameters were tuned using random search (60 points) and a 3-fold-CV procedure was used for error estimation. \texttt{SVR}'s MSE is significantly larger than \texttt{GAM}'s one, and still, there are several regions of the input space where \texttt{GAM} is outperformed (in light colors). Panels (b,c)  depict the regression surface of two models learned using tree-based Gradient Boosting machines (\texttt{GB}) and Random Forests (\texttt{RF}), respectively, with 100 trees and default hyperparameter settings. To show their sensitivity to \emph{target} outliers, we artificially imputed one extremely high value (in black) in the target of one single example (where the value is already expected to be maximum). In Panels (d,e), we analyze the same effects with two stacking approaches using the models fitted in (a,b,c) as base learners: Linear Stacking (\texttt{LS}) in (d) and Dynamic Selection (\texttt{DS}) with \texttt{kNN} in (e). Please note how deformed the regression surfaces (in gray) are in all settings (b-d). Panel (f) depicts the original surface. Best viewed in color.}
	\label{fig:intro_regression}
\end{figure*}

We can group different approaches for model integration in three clusters \cite{todorovski03}: (a) voting (e.g. bagging \cite{breiman96}), (b) cascading \cite{Gama2000} and (c) stacking \cite{wolpert1992}. In voting, the outputs of the ensemble is a (weighted) average of outputs of the base models. Cascading iteratively combines the outputs of the base experts by including them, one of a time, as yet another feature in the training set. Stacking learns a meta-model that combines the outputs of all the base models. All these approaches have advantages and shortcomings. Voting relies on base models to have some complementary expertise\footnote{Some base models perform reasonably well in some subregion of the feature space, while other base models perform well in other regions.}, which is an assumption that is rarely true in practice (e.g. check Fig.~\ref{fig:intro_regression}-(b,c)). On the other hand, cascading typically results in complex and time-consuming to put in practice, since it involves training of several models in a sequential fashion.

Stacking relies on the power of the meta-learning algorithm to approximate $\hat{F}$. It is possible to group stacking approaches in two types: parametric and non-parametric. The first (and more commonly used \cite{kaggle2018a}) is based on assuming apriori a (typically linear) functional form for $\hat{F}$, while its coefficients are either learned or estimated somehow \cite{breiman1996stacked}. The second follows a strict meta-learning approach \cite{brazdil2008}, where a meta-model for $\hat{F}$ is learned in a non-parametric fashion by relating the characteristics of problems (i.e. properties of the training data) with the performance of the experts. Notable approaches include instance-based learning \cite{tsymbal2006} and decision trees \cite{todorovski03}. However, as for many other problems in supervised learning, novel approaches for model integration in ensemble learning are primarily designed for classification and, if at all, adapted later on for regression \cite{todorovski03,tsymbal2006,mendes2012}. While such adaptation may be trivial in many cases, it is important to note that regression poses distinct challenges.

Formally, we may formulate a classical regression problem as the problem of learning a function
\begin{equation}
\hat{f_{\theta}}: x_i \rightarrow \mathbb{R} \quad \mathrm{such\ that}\quad \hat{f}(x_i;\theta) \simeq f(x_i)=y_i, \forall x_i \in X, y_i \in Y
\end{equation}
where $f(x_i)$ denotes the true unknown function which is generating the samples' target variable values, and $\hat{f}(x_i;\theta)=\hat{y}_i$ denotes an approximation dependent on the feature vector $x_i$ and an unknown (hyper)parameter vector $\theta \in \mathbb{R}^n$. One of the key differences between regression and classification is that for regression the range 
 of $f$ is apriori undefined and potentially infinite. This issue raises practical hazards for applying many of the widely used supervised learning algorithms, since some of them cannot predict outside of the target range of their training set values (e.g. Generalized Additive Models (\texttt{GAM}) \cite{hastie1987} or \texttt{CART} \cite{breiman1984}). Another major issue in regression problems are \textbf{outliers}. In classification, one can observe either \emph{feature} or \emph{concept} outliers (i.e. outliers in $p(x)$ and $p(y|x)$), while in regression one can also observe \emph{target} outliers (in $p(y)$). Given that the true target domain is unknown, these outliers may be very difficult or even impossible to handle with common preprocessing techniques (e.g. Tukey's boxplot or one-class SVM \cite{chandola09}). Fig. \ref{fig:intro_regression} illustrates these issues in practice on a synthetic example with different regression algorithms. Although the idea of training different experts in parallel to subsequently combine them seems theoretically attractive, the abovementioned issues make it hard to use practice, especially for regression. In this context, stacking is regarded to be a complex \emph{art} of finding the right combination of data preprocessing, model generation/pruning/integration and post-processing approaches for a given problem.

In this paper, we introduce \texttt{MetaBags}, a novel, practically useful stacking framework for regression. \texttt{MetaBags} is a powerful meta-learning algorithm that learns a set of meta-decision trees designed to select one expert for each query thus reducing inductive bias. These trees are learned using different types of meta-features specially created for this purpose on data bootstrap samples, whereas the final meta-model output is the average of the outputs of the experts selected by each meta-decision tree for a given query. Our contributions are three fold:
\begin{enumerate}
\item A novel meta-learning algorithm to perform non-parametric stacking for regression problems with \textbf{minimum user expertise requirements}.
\item An approach for turning the traditional overfitting tendency of stacking into an advantage through the usage of \textbf{bagging at the meta-level}.
\item A novel set of \textbf{local landmarking meta-features} that characterize the learning process in feature subspaces and enable model integration for regression problems.
\end{enumerate}
In the remainder of this paper, we describe the proposed approach, after discussing related work. An exhaustive experimental evaluation of its efficiency and scalability in practice. This evaluation employs 17 regression datasets (including one real-world application) and compares our approach to existing ones.

	\section{Related Work}
Since its first appearance, meta-learning has been defined in multiple ways that focus on different aspects such as collected experience, domain of application, interaction between learners and the knowledge about the learners \cite{lemke2015}. Brazdil et al. \cite{brazdil2008} define meta-learning as the learning that deals with both types of bias, declarative and procedural. The declarative bias is imposed by the hypothesis space form which a base learner chooses a model, whereas the procedural bias defines how different hypotheses should be ordered/preferred.
In a recent survey, Lemke et al. \cite{lemke2015} characterize meta-learning as the learning that constitutes three essential aspects: (i) the adaptation with experience, (ii) the consideration of meta-knowledge of the data set (to be learned from) and (iii) the integration of meta-knowledge from various domains. Under this definition, both ensemble methods bagging \cite{breiman96} and boosting \cite{Freund1997} do not qualify as meta-learners, since the base learners in bagging are trained independently of each other, and in boosting, no meta-knowledge from different domains is used when combining decisions from the base learners. Using the same argument, stacking \cite{wolpert1992} and cascading \cite{Gama2000} cannot be definitely considered as meta-learners \cite{lemke2015}. 

Algorithm recommendation, in the context of meta-learning, aims to propose the type of learner that best fits a specific problem. This recommendation can be performed after considering both the learner's performance and the characteristics of the problem \cite{lemke2015}. Both aforementioned aspects qualify as meta-features that assist in deciding which learner could perform best on a specific problem. Meta-features are often categorized into three different classes of meta-features \cite{brazdil2008}: (i) meta-features of the dataset describing its statistical properties such as the number of classes and attributes, the ratio of target classes, the correlation between the attributes themselves, and between the attributes and the target concept, (ii) model-based meta-features that can be extracted from models learned on the target dataset, such as the number of support vectors when applying SVM, or the number of rules when learning a system of rules, and (iii) landmarkers, which constitute the generalization performance of diverse set of learners on the target dataset in order to gain insights into which type of learners fits best to which regions/subspaces of the studied problem. Traditionally, landmarkers have been mostly proposed in a classification context \cite{pfahringer2000,brazdil2008}. A notorious exception is proposed by Feurer et al. \cite{feurer2015}. The authors use meta-learning to generate prior knowledge to feed a bayesian optimization procedure in order to find the best sequence of algorithms to address a predefined set of tasks on either classification and regression pipelines. However, the original paper describing its meta-learning procedure \cite{feurer2015a} is focused mainly on classification. 

The dynamic approach of ensemble integration \cite{mendes2012} postpones the integration step till prediction time so that the models used for prediction are chosen dynamically, depending on the query to be classified. Merz 
\cite{Merz1996} applies dynamic selection (\texttt{DS}) locally by selecting models that have good performance in the neighborhood of the observed query. This can be seen as an integration approach that considers type-(iii) landmarkers. Tsymbal et al. \cite{tsymbal2006} show how \texttt{DS} for random forests decreases the bias while keeping the variance unchanged.

In a classification setting, Todorovski and D\v{z}eroski \cite{todorovski03} combine a set of base classifiers by using meta-decision trees which in a leaf node give a recommendation of a specific classifier to be used for instances reaching that leaf node. Meta-decision trees (\texttt{MDT}) are learned by stacking and use the confidence of the base classifiers as meta-features. These can be viewed as landmarks that characterizes the learner, the data used for learning and the example that needs to be classified. Most of the suggested meta-features, as well as the proposed impurity function used for learning \texttt{MDT} are applicable to classification problems only.

\texttt{MetaBags} can be seen as a generalization of \texttt{DS} \cite{Merz1996,tsymbal2006} that uses meta-features instead. Moreover, we considerably reduce \texttt{DS}  runtime complexity (generically, $\mathcal{O}(N)$ in test time, even with the most modern search heuristics \cite{beygelzimer2006}), as well as the user-expertise requirements to develop a proper metric for each problem.  Finally, the novel type of local landmarking meta-features characterize the local learning process - aiming to avoid overfitting when a particular input subspace is not well covered in the training set.
	\section{Methodology}
\begin{figure*}[!t]
	\centering
	\includegraphics[trim=2.2cm 16.5cm 0.9cm 0.2cm, width=\linewidth,clip=TRUE]{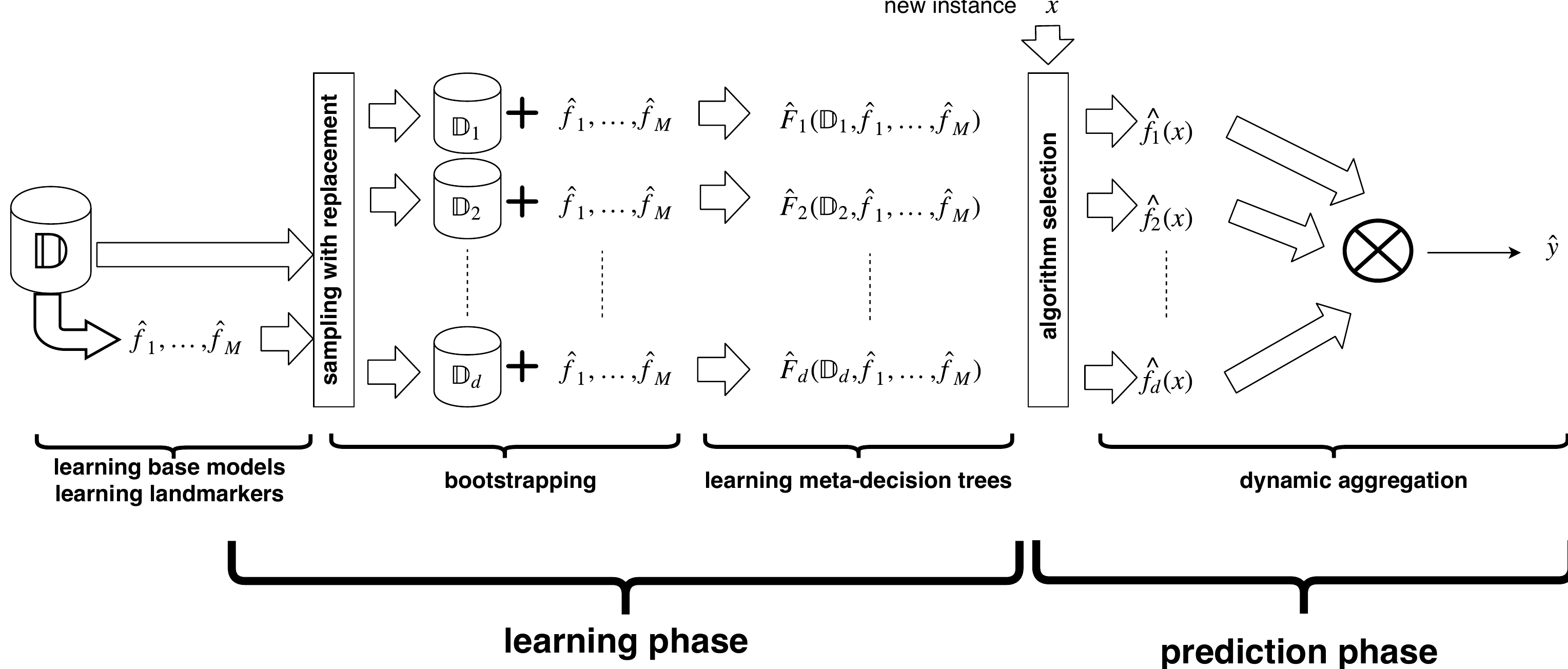}
	\caption{\texttt{MetaBags}: the learning/induction and prediction phases.}
	\label{fig:Metaags_Process}
\end{figure*}

This Section introduces \texttt{MetaBags} and its three basic components: (1) we firstly describe a novel algorithm to learn a decision tree that picks one expert among all available ones to address a particular query in a supervised learning context; (2) then, we depict the integration of base models at the meta-level with bagging to form the final predictor $\hat{F}$; (3) Finally, the meta-features used by \texttt{MetaBags} are detailed. An overview of the whole method is presented in Fig.\ \ref{fig:Metaags_Process}.
\subsection{Meta-Decision Tree for Regression}
\subsubsection{Problem Setting.}
In traditional stacking, $\hat{F}$ just depends on the base models $\hat{f}_i$. In practice, as stronger models may outperform weaker ones (c.f. Fig. \ref{fig:intro_regression}-(a)), and get assigned very high coefficients(assuming we combine base models with a linear meta-model). In turn, weaker models may obtain near-zero coefficients, since those are learned by taking into account the same training set where the base models were learned. This can easily leads to over-fitting if whenever a careful model generation does not take place beforehand (c.f. Fig. \ref{fig:intro_regression}-(d,e)). However, even a model that is weak in the whole input space may be strong in some subregion. In our approach we rely on classic tree-based isothetic boundaries to identify \textit{contexts} (e.g. subregions of the input space) where some models may outperform others, and by using only strong experts within each context, we improve the overall model. 

Let the dataset $\mathbb{D}$ be defined as $(x_i,y_i) \in \mathbb{D} \subset \mathbb{R}^n \times \mathbb{R}: i=\{1, \ldots, N\}$ and generated by an unknown function $f(x)=y$, where $n$ is the number of features of an instance $x$, and $y$ denotes a numerical response. Let $\hat{f}_j (x): j=\{1..M\}$ be a set of $M$ base models (\emph{experts}) learned using one or more base learning methods over $\mathbb{D}$. Let $\mathcal{L}$ denote a loss function of interest decomposable in independent bias/variance components (e.g.\ $L2$-loss). For each instance $x_i$, let $\{z_{i,1},\ldots,z_{i,Q}\}$ be the set of meta-features generated for that instance.

Starting from the definition of a decision tree for supervised learning, introduced in \texttt{CART} \cite{breiman1984}, we aim to build a \textit{classification} tree that, for a given instance $x$ and its supporting meta-features $\{z_{1},\ldots,z_{Q}\}$, dynamically selects the expert that should be chosen for prediction, i.e., $\hat{F}(x,z_{1},\ldots,z_{Q};\hat{f_1},\ldots,\hat{f_M})=\hat{f_j}(x)$. 
As for the tree induction procedure, like \texttt{CART}, we aim, at each node, at finding the feature $z_j$ and the spiting point $z_j^t$ that leads to the maximum reduction of impurity.
For the internal node $p$ with the set of examples $\mathbb{D}_p \in \mathbb{D}$ that reaches $p$, the splitting point $z_j^t$ splits the node $p$ into the leaves $p_l$ and $p_r$ with the sets $\mathbb{D}_{p_l} = \{x_i \in \mathbb{D}_p | z_{ij} \leq z_{j}^t\}$ and 
$\mathbb{D}_{p_r} = \{x_i \in \mathbb{D}_p | z_{ij} > z_{j}^t\}$, respectively. This can be formulated by the following optimization problem at each node:
\begin{align}
\label{eq_meta_decision_tree_base}
& \underset{z_j^t}{\arg \max} \quad \omega(z_j^t)  \\
\label{eq_meta_decision_tree_base_2}
 \text{s.t.} \qquad \omega(z_j^t) & = [I(p)-P_lI(p_l)-P_rI(p_l)]
\end{align}
where $P_l, P_r$ denote the probability of each branch to be selected, while $I(t)$ denotes the so-called \textit{impurity function}. In traditional classification problems, the functions applied here aim to minimize the entropy of the target variable. Hereby, we propose a new impurity function for this purpose denoted as \underline{M}aximum \underline{B}ias \underline{R}eduction. It goes as follows:
\begin{equation}
\label{eq_impurity}
I(t) = MBR(t)=\underset{j \in \{1...M\}}{\max} - E\Big[B\big(\mathcal{L}(t,\hat{f}_j)\big)^2\Big]
\end{equation}
where $B(\mathcal{L})$ denotes the inductive bias-decomposition of the loss $\mathcal{L}$.
\subsubsection{Optimization.} To solve the problem of Eq.~(\ref{eq_meta_decision_tree_base}), we address three issues: (i) splitting criterion/meta-feature, (ii) splitting point and (iii) stopping criterion. To select the splitting criterion, we start by constructing two auxiliary equally-sized matrices $a \in \mathbb{R}^{Q \times \phi}$ and $b: z_{i_{\min}} \leq b_{i,j} \leq z_{i_{\max}}, \forall i,j$, where $\phi \in \mathbb{N}$ denotes a user-defined hyperparameter. Then, the matrices are populated with candidate values by elaborating over the eqs. (\ref{eq_meta_decision_tree_base},\ref{eq_meta_decision_tree_base_2},\ref{eq_impurity}) as 
\begin{equation}
\label{eq_matrix1}
a_{i,j}=\omega(b_{i,j}), \quad b_{i,j} \sim U(z_{i_{\min}},z_{i_{\max}}) ,
\end{equation}
where $b_{i,j}$ is the $j$th splitting criterion for the $i$th meta feature.

At first we find the splitting criteria $\tau$ such that 
\begin{equation}
\tau = \underset{i\mbox{ }\in\mbox{ }\{1..Q\}}{\arg\min}\mbox{ }a_{i,j}, \quad \forall j \in \{1..\phi\} .
\end{equation}
Secondly, we need to find the optimal splitting point according to the $z_{\tau}$ criteria. A natural choice for this problem is a simplified Golden-section search algorithm \cite{kiefer1953}: it is simple, scales reasonably and can be trivially initialized by the values in the $\tau$th row in the matrix $b$ row. The only constraint is the maximum number of iterations $\rho \in \mathbb{N}$, which is an user-defined hyperparameter.
Thirdly, (iii) the stopping criteria to constraint eq. (\ref{eq_meta_decision_tree_base}). Here, like \texttt{CART}, we propose to fully grown trees. Therefore, it goes as follows:
\begin{equation}
\omega\big(z_\tau^t\big) \geq \epsilon \mbox{ } \wedge \mbox{ } |\mathbb{D}_p| \geq \upsilon: \epsilon \in \mathbb{R}^{+}, \upsilon \in \mathbb{N}
\end{equation}
where $\epsilon, \upsilon$ are user-defined hyperparameters. 

The pseudocode of this algorithm is presented in Algorithm 1.
\subsection{Bagging at Meta-Level: Why and How?}
Bagging \cite{breiman96} is a popular ensemble learning technique. It consists of forming multiple $d$ replicate datasets $\mathbb{D}^{(B)} \subset \mathbb{D}$ by drawning $s << N$ examples from $\mathbb{D}$ at random, but with replacement, forming bootstrap samples. Next, $d$ base models $\varphi(x_i,\mathbb{D}^{(B)})$ are learned with a selected method on each  $\mathbb{D}^{(B)}$, and the final prediction $\varphi_A(x_i)$ is obtained by averaging the predictions of all $d$ base models. As Breiman demonstrates in Section 4 of \cite{breiman96}, the amount of expected improvement of the aggregated prediction $\varphi_A(x_i)$ depends on the gap between the terms of the following inequality:
\begin{equation}
\label{eq_bagging}
E\big[\mathcal{L}\big(\varphi(x_i,\mathbb{D}^{(B)})\big)\big]^2 \leq E\big[\mathcal{L}\big(\varphi(x_i,\mathbb{D}^{(B)})\big)^2\big].
\end{equation}
In our case, $\varphi(x_i,\mathbb{D}^{(B)})$ is given by the $\hat{f}_j(x_i)$ selected by each meta-decision tree induced in each $\mathbb{D}^{(B)}$. By design, the procedure to learn this specific meta-decision tree is likely to overfit its training set, since all the decisions envisage reduction of inductive bias alone. However, when used in a bagging context, this turns to be an advantage because it causes a  instability of $\varphi$ - as each tree may be selecting different predictors to each instance $x_i$. This is more likely as more as the dominant regions (i.e. meta-features subspaces) of each expert $\hat{f}$ on our meta-decision space (c.f. Fig. \ref{fig:intro_regression}-(a)) are equally-sized. 
\subsection{Meta-Features}
\texttt{MetaBags} is fed with three types of meta-features:  (a) base , (b) performance-related and (c) Local Landmarking. These types are briefly explained below, as well as their connection with the State-of-the-Art in the area.
\subsubsection{(a) Base features}
Following \cite{todorovski03}, we propose to include all base features also as meta-features. This aims to stimulate a higher inequality in Eq. (\ref{eq_bagging}) due to the increase of inductive variance of each individual meta-predictor. 

\subsubsection{(b) Performance-related features.}
This type of meta-features describe the performance of specific learning algorithms in particular learning contexts on the same dataset. Besides the base learning algorithms, we also propose the usage of landmarkers. Landmarkers are ML algorithms that are computationally relatively \textit{cheap} to run either in a train or test setting \cite{pfahringer2000}. The resulting models aim to characterize the learning task (e.g. is the regression curve linear?). To the authors best knowledge, so far, all proposed landmarkers and consequent meta-features have been primarily designed for classical meta-learning applications to classification problems \cite{pfahringer2000,brazdil2008}, whereas we focus on model integration for regression. We use the following learning algorithms as landmarkers: \texttt{LASSO} \cite{tibshirani1996}, \texttt{1NN} \cite{cover1967}, \texttt{MARS} \cite{friedman1991} and \texttt{CART} \cite{breiman1984}.

To generate this set of meta-features, we start by creating one landmarking model per each available method over the entire training set. Then, we design a \textit{small} artificial neighborhood of size $\psi$ of each training example $x_i$ as ${{X'}_i}=\{{x'}_{i,1}, {x'}_{i,2}..{x'}_{i,\psi}\}$ by perturbing $x_i$ with \textit{gaussian noise} as follows:
\begin{equation}
{x'}_{i,j} = x_{i,j} + \xi: \xi \sim \mathcal{N}_n(0,1), \forall j \in \{1,..,\psi\} 
\end{equation}
where $\psi,$ is an user-defined hyperparameter. Then, we obtain outputs of each expert as well as of each landmarker given $X'_i$. The used meta-features are then descriptive statistics of the models' outputs: mean, stdev., 1st/3rd quantile. 
\subsubsection{(c) Local landmarking features.}
In the original landmarking paper, Pfahringer \textit{et al.} \cite{pfahringer2000} highlight the importance on ensuring that our pool of landmarkers is diverse enough in terms of the different types of inductive bias that they employ, and the consequent relationship that this may have with the base learners performance. However, when observing performance at neighborhood-level rather than on the task/dataset level, the low performance and/or high inductive bias may have different causes (e.g., inadequate data preprocessing techniques, low support/coverage of a particular subregion of the input space, etc.). These causes, although having a similar effect, may originate in different types of deficiencies of the model (e.g. low support of leaf nodes or high variance of the examples used to make the predictions in decision trees).

Hereby, we introduce a novel type of landmarking meta-features denoted \textbf{local landmarking}. Local landmarking meta-features are designed to characterize the landmarkers/models within the particular input subregion. More than finding a correspondence between the performance of landmarkers and base models, we aim to extract the knowledge that the landmarkers have learned about a particular input neighborhood. In addition to the prediction of each landmarker for a given test example, we compute the following characteristics:
\begin{itemize}
\item \texttt{CART}: depth of the leaf which makes the prediction; number of examples in that leaf and variance of these examples;
\item \texttt{MARS}: width and mass of the interval in which a test example falls, as well as its distance to the nearest edge;
\item \texttt{1NN}: absolute distance to the nearest neighbor.
\end{itemize}

	\section{Experiments and Results}

Empirical evaluation aims to answer the following four research questions:
\begin{enumerate}[label=(\textbf{Q\arabic*}),align=left, leftmargin=*]
	\item Does \texttt{MetaBags} systematically outperform its base models in practice?
	\item Does \texttt{MetaBags} outperform other model integration procedures?
    \item Do the local landmarking meta-features improve \texttt{MetaBags} performance?
	\item Does \texttt{MetaBags} scale on large-scale and/or high-dimensional data?
\end{enumerate}
In the reminder of this Section we present the datasets used for evaluation, evaluation methodology and results.

\subsection{Regression Tasks}
We used a total of 17 benchmarking datasets to evaluate \texttt{MetaBags}. They are summarized in Table \ref{table:usedDataSets}.In addition, we include 4 proprietary datasets addressing a particular real-world application: public transportation. One of its most common research problems is travel time prediction (TTP). Attaining better bus TTPs can have significant consequences for passenger delays, operator's performance fines and the efficiency of its resource allocation. The work in \cite{Hassan2016} in TTP uses features such as scheduled departure time, vehicle type and/ or driver's meta-data. This type of data is known to be particularly noisy due to failures in the data collecting procedures, either hardware or human-related, which in turn often lead to several issues such as missing and/or unreliable data measures, as well as several types of outliers \cite{moreira-matias2015}.

Here, we evaluate \texttt{MetaBags} in a similar setting of \cite{Hassan2016}, i.e. by using their four datasets and the original preprocessing. This case study is an undisclosed large urban bus operator in Sweden (BOS). We collected data on four high-frequency routes/datasets R11/R12/R21/R22. These datasets cover a time period of six months.
\subsection{Evaluation Methodology}
Hereby, we describe the empirical methodology designed to answer (Q1-Q4), including the hyperparameter settings of \texttt{MetaBags} and the algorithms selected for comparing the different experiments.
\subsubsection{Hyperparameter settings.}
Like many other decision tree-based algorithms, \texttt{MetaBags} is expected to be robust to its hyperparameter settings. Table \ref{tab:hyperparameters} presents the hyperparameters settings used in the empirical evaluation (a sensible default). If any, $s$ and $d$ can be regarded as more sensitive parameters. Their value ranges are recommended to be $0 < s << 1$ and $100 \leq d << 2000$.

\subsubsection{Testing scenarios and comparison algorithms.}
We put in place two testing scenarios: A and B. In scenario A, we evaluate the generalization error of \texttt{MetaBags} with 5-fold cross validation (CV) with 3 repetitions. As base learners, we use four popular regression algorithms: Support Vector Regression (\texttt{SVR})\cite{drucker97}, Projection Pursuit Regression (\texttt{PPR})\cite{friedman81}, Random Forest \texttt{RF} \cite{breiman01} and Gradient Boosting \texttt{GB} \cite{friedman2001}. The first two are popular methods in the chosen application domain \cite{Hassan2016}, while the latter are popular voting-based ensemble methods for regression \cite{kaggle2018a}. The base models had their hyperparameter values tuned with random search/3-fold CV (and 60 evaluation points). We used the implementations in the R package \bera{[caret]} for both the landmarkers and the base learners \footnote{Experimental source code will be made publicly available.}. We compare our method to the following ensemble approaches: Linear Stacking \texttt{LS} \cite{breiman1996stacked}, Dynamic Selection \texttt{DS} with \texttt{kNN} \cite{tsymbal2006,Merz1996}, and the best individual model. All methods used $l2$-loss as $\mathcal{L}$.

In scenario B, we extend the artificial dataset used in Fig.\ \ref{fig:intro_regression} to assess the computational runtime scalability of the decision tree induction process of \texttt{MetaReg} (using a CART-based implementation) in terms of number of examples and attributes. In this context, we compare our method's training stage to Linear Regression (used for \texttt{LS}) and \texttt{kNN} in terms of time to build k-d tree (\texttt{DS}). Additionally, we also benchmarked \texttt{C4.5}  (which was used in \texttt{MDT} \cite{todorovski03}). For the latter, we discretized the target variable using the four quantiles.
\subsection{Results}
Table \ref{results_base} presents the performance results of \texttt{MetaBags} against comparison algorithms: the base learners; SoA in model integration such as stacking with a linear model \texttt{LS} and kNN, i.e. \texttt{DS}, as well as the best base model selected using 3-CV i.e. \texttt{Best}; finally, we also included two variants of \texttt{MetaBags}: \texttt{MetaReg} -- a singular decision tree, \texttt{MBwLM} -- \texttt{MetaBags} without the novel landmarking features. Results are reported in terms of RMSE, as well as of statistical significance (using the using the two-sample \textit{t}-test with the significance level $\alpha = 0.05$). Finally, Fig. \ref{fig:summary_results} summarizes those results in terms of percentual improvements, while Fig. \ref{fig:scalability} depicts our empirical scalability study.
\afterpage{
\begin{table*}[!t]
	\begin{center}
    \caption{Datasets summary. Fields denote number of \underline{\#ATT}ributes and \underline{\#INS}tances, the Range of the Target variable, the number of \underline{\#T}arget \underline{O}utliers using Tukey's boxplot(ranges=1.5,3), their \underline{ORI}gin, as well as its \underline{TYP}e (\textbf{P}roprietary/\textbf{O}pen) and \underline{C}ollection \underline{P}rocess (\textbf{R}eal/\textbf{S}imulated/\textbf{A}rtificial).}
		\scalebox{0.8}{\begin{tabular}{@{\extracolsep{\fill}} l l l l l l l l l}
			\toprule
            \label{table:usedDataSets}
			&\multicolumn{5}{l}{Properties}
			&\multicolumn{3}{l}{Source and Type}\\
			\cline{2-6}\cline{7-9}
			&\rotatebox[origin=l]{0}{\#ATT}
			&\rotatebox[origin=l]{0}{\#INS}
			&\rotatebox[origin=l]{0}{RT}
            &\rotatebox[origin=l]{0}{\#TO(1.5)}
            &\rotatebox[origin=l]{0}{\#TO(3.0)}
			&\rotatebox[origin=l]{0}{ORI}
            &\rotatebox[origin=l]{0}{TYP }
			&\rotatebox[origin=l]{0}{CP}
			\\	
			\hline
			R11&12&17953&[1306,10520]&66&9&BOS&P&R\\
			R12&12&16353&[1507,9338]&154&6&BOS&P&R\\
			R21&12&16280&[1434,6764]&341&27&BOS&P&R\\
			R22&12&16353&[884,6917]&146&10&BOS&P&R\\
			\hline
			Cal. housing&8&20460&[14999,500001]&1071&0&StatLib&O&R\\
			Concrete&8&1030&[2332,82599]&4&0&UCI&O&R\\
			2Dplanes&10&40768&[-999.709,999.961]&4&0&dcc.fc.up.pt\footnote{\url{https://www.dcc.fc.up.pt/~ltorgo/Regression/2dplanes.html}}&O&A\\
			Delta Ailerons&6&7129&[-0.0021,0,0022]&107&12&dcc.fc.up.pt\footnote{\url{https://www.dcc.fc.up.pt/~ltorgo/Regression/delta_ailerons.html}}&O&R\\
			Elevators&18&16559&[0.012,0,078]&842&344&dcc.fc.up.pt\footnote{\url{https://www.dcc.fc.up.pt/~ltorgo/Regression/elevators.html}}&O&R\\
			Parkinsons Tele.&26&5875&[0.022,0732]&206&17&UCI&O&R\\
			Physicochemical&9&45730&[15.228,55.3009]&0&0&UCI&O&R\\
			Pole&48&15000&[0,100]&0&0&dcc.fc.up.pt\footnote{\url{https://www.dcc.fc.up.pt/~ltorgo/Regression/pole.html}}&O&R\\
			Puma32H&32&8192&[-0.085173,0.088266]&56&0&DELVE&O&R\\
       		Red wine quality&12&1599&[3,8]&28&0&UCI&O&R\\
            White wine quality&12&4898&[3,9]&200&0&UCI&O&R\\
		    \textbf{Computer Activity }&&&&&& \\
			\hspace{6mm}CPU-small&12&8192&[0,99]&430&294&DELVE&O&R\\
			\hspace{6mm}CPU-activity&21&8192&[0,99]&430&294&DELVE&O&R\\
			\hline
		\end{tabular}}
	\end{center}		
\end{table*}
\begin{table*}[!t]

	\begin{center} 
		\caption{Hyperparameter settings used in \texttt{MetaBags}.}
		\label{tab:hyperparameters}
		\scalebox{0.8}{
			\begin{tabular}{c | c | c }
				\toprule
				\hline
				{\scalebox{1.1}{\textbf{}}} & {\scalebox{1.1}{\textbf{Value}}} & {\scalebox{1.1}{\textbf{Description}}} \\
				\hline
				$\phi$ & $10$ & \begin{tabular}{@{}l@{}}number of random partitions performed on each meta-feature;\end{tabular} \\
                $\rho$ & $3$ & \begin{tabular}{@{}l@{}}max. number of iterations on finding optimal splitting point;\end{tabular} \\
                 $\epsilon$ & $|I(t_p)\cdot10^{-2}|$ & \begin{tabular}{@{}l@{}}min. abs. bias reduction to perform split;\end{tabular} \\
                 $\upsilon$ & $N\cdot10^{-2}$ & \begin{tabular}{@{}l@{}}min. examples in node to perform split;\end{tabular} \\
                 $\psi$ & $100$ & \begin{tabular}{@{}l@{}}size of the artificial neighborhood generated to compute meta-features;\end{tabular}  \\
                 $s$ & $10\%$ & \begin{tabular}{@{}l@{}}percentage of examples usage to generate the bootstraps;\end{tabular}  \\
                 $d$ & $300$ & \begin{tabular}{@{}l@{}}number of generated meta-decision trees;\end{tabular}  \\   
				\hline
				\bottomrule
			\end{tabular}
		}
	\end{center}	
\end{table*}
\begin{figure*}[!t]
	\centering
    \includegraphics[trim=0.21cm 0.7cm 0.13cm 0.45cm, width=0.8\linewidth,clip=TRUE]{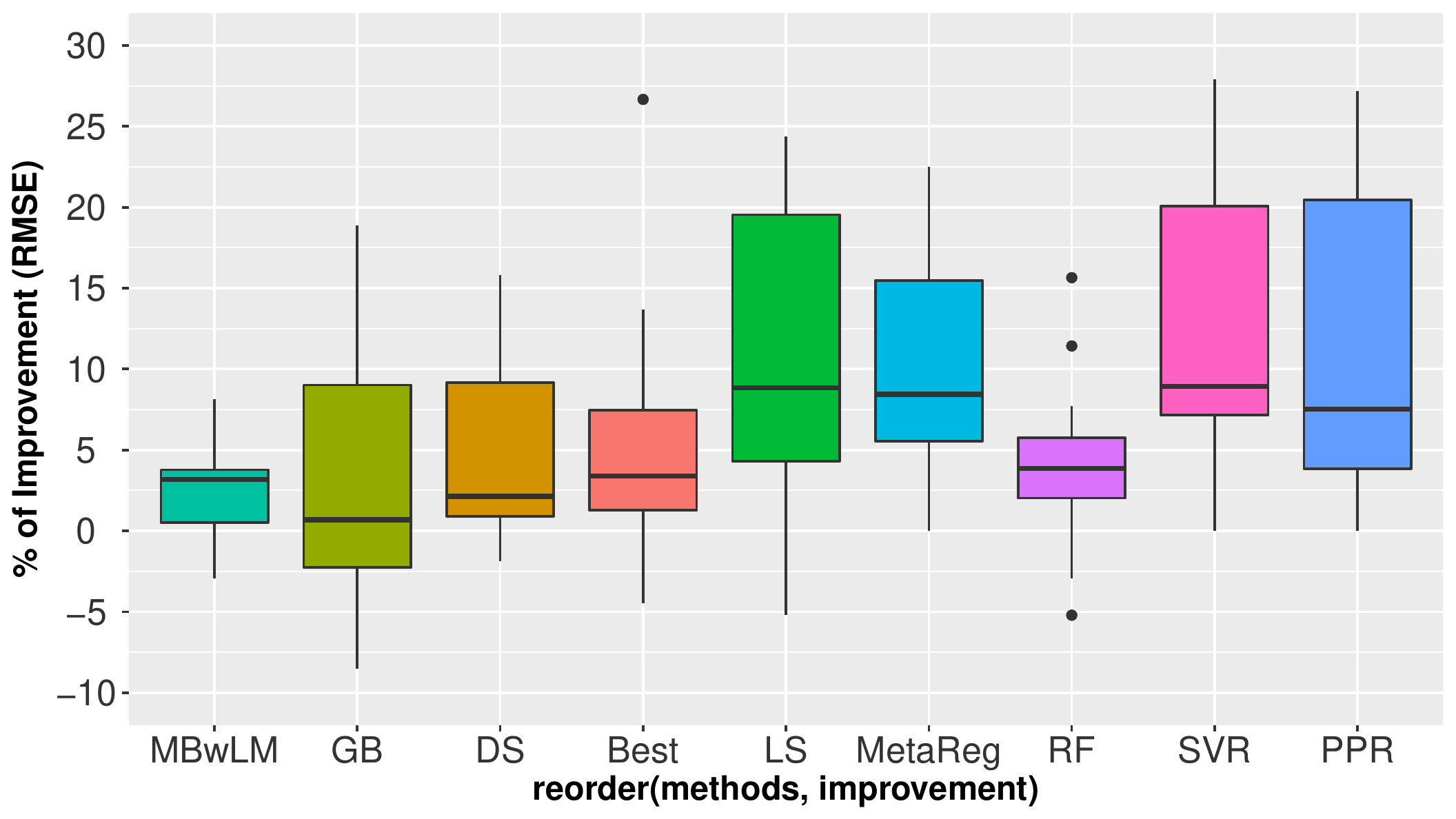}
	\caption{Summary results of \texttt{MetaBags} using the percentage of improvement over its competitors. Note the consistently positive mean over all methods.}
	\label{fig:summary_results}
\end{figure*}}
\afterpage{
{\fontsize{5}{5}\selectfont
\begin{algorithm*}[!t]
	\label{alg:InduceMetaDecisionTreeRegression}
	\caption{InduceMetaDecisionTreeRegression}
    \scriptsize
	\KwIn{
		$p$: root (or current internal node) of the meta decision tree.\\
		$\mathbb{D}_p \subset \mathbb{D}_i$: the subset of examples that reach the root (or internal node) $p$, $\mathbb{D}_i$ is a bootstrap sampled with replacement from $\mathbb{D}$.\\
		$\{\hat{f}_j | j \in \{1..M\} \}$: the set of base models.\\
		$\{z_{u,v} | x_u \in \mathbb{D}_p \wedge v \in \{1,\ldots,Q\}\}$: the set of meta features for each instance in $\mathbb{D}_p$.
	}
	\tcc{check that the current node has the minimum number of supporting instances}	
	\If{$|\mathbb{D}_p| \geq \upsilon$}
	{
		\Return		
	}		
	\tcc{create the matrices $A$ and $B$}	
	$B = [b_{i,j}]\in \mathbb{R}^{Q \times \phi} \quad \text{s.t.} \quad b_{i,j} \sim U(z_{i_{\min}},z_{i_{\max}}) $ \\
	$A = [a_{i,j}]\in \mathbb{R}^{Q \times \phi} \quad \text{s.t.} \quad
	a_{i,j} = \omega(b_{i,j})  = [I(p)-P_lI(p_l)-P_rI(p_l)]$\\
	\tcc{find the splitting criteria $\tau$}
	$\tau = \underset{i\mbox{ }\in\mbox{ }\{1..Q\}}{\arg\min}\mbox{ }a_{i,j}, \quad \text{for all} \quad j \in \{1,\ldots,\phi\}$\\
	\tcc{fine tune found splitting point on the $\tau$th criteria}
	apply Golden-section search algorithm to find $z_\tau^t \in \{z_{\tau_{\min}},z_{\tau_{\max}}\}$\\
	\If{$\omega\big(z_\tau^t\big) \geq \epsilon$}
	{
		\tcc{create the right and left leaf nodes, $p_l$ and $p_r$}
		$\mathbb{D}_{p_l} = \{x_i \in \mathbb{D}_p | z_{i\tau} \leq z_{\tau}^t\}$\\
		$\mathbb{D}_{p_r} = \{x_i \in \mathbb{D}_p | z_{i\tau} > z_{\tau}^t\}$\\
		$\text{InduceMetaDecisionTree}(p_l,\mathbb{D}_{p_l},\{\hat{f}_j\},\{z_{u,v}\})$\\
		$\text{InduceMetaDecisionTree}(p_r,\mathbb{D}_{p_r},\{\hat{f}_j\},\{z_{u,v}\})$\\
	}
	\Return	
    
\end{algorithm*}
}
%
\begin{figure*}[!t]
	\centering
    \includegraphics[trim=0.5cm 0.4cm 0.2cm 0.0cm, width=0.5\textwidth,clip=TRUE]{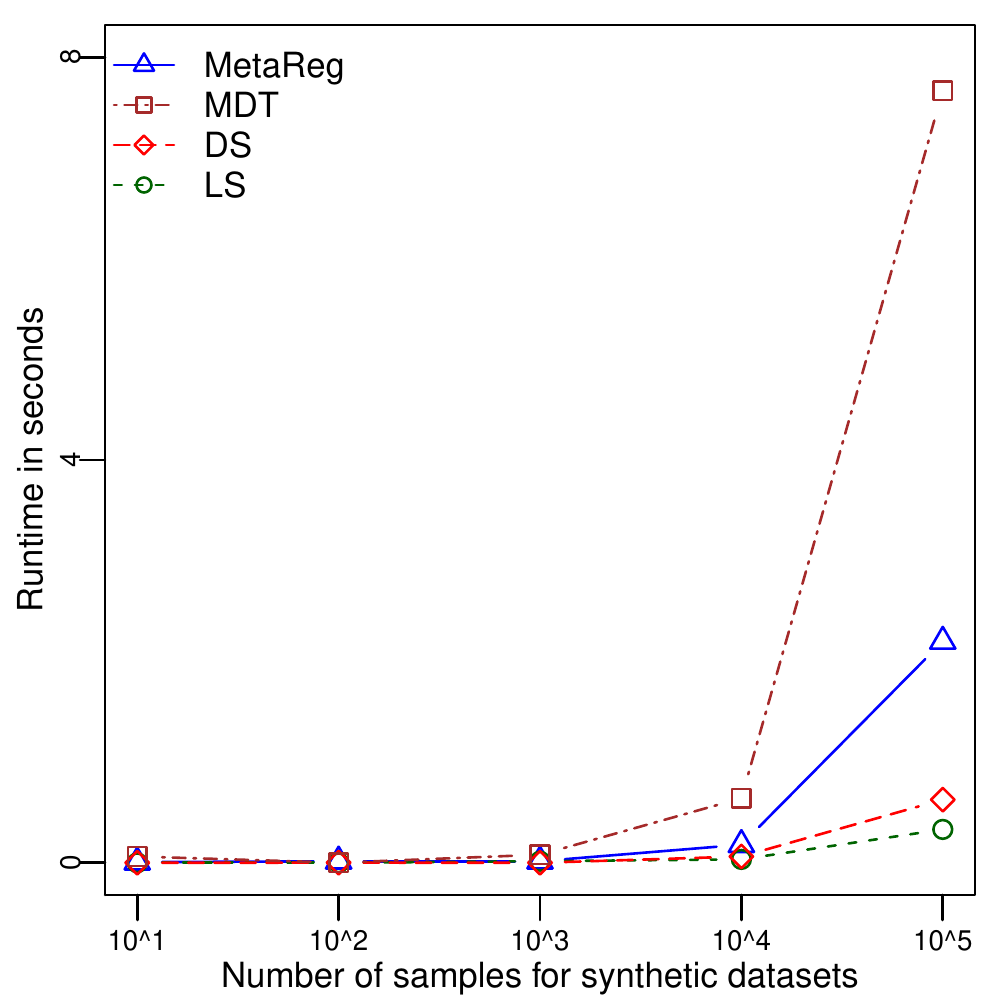}
        \includegraphics[trim=0.5cm 0.4cm 0.2cm 0.0cm, width=0.5\textwidth,clip=TRUE]{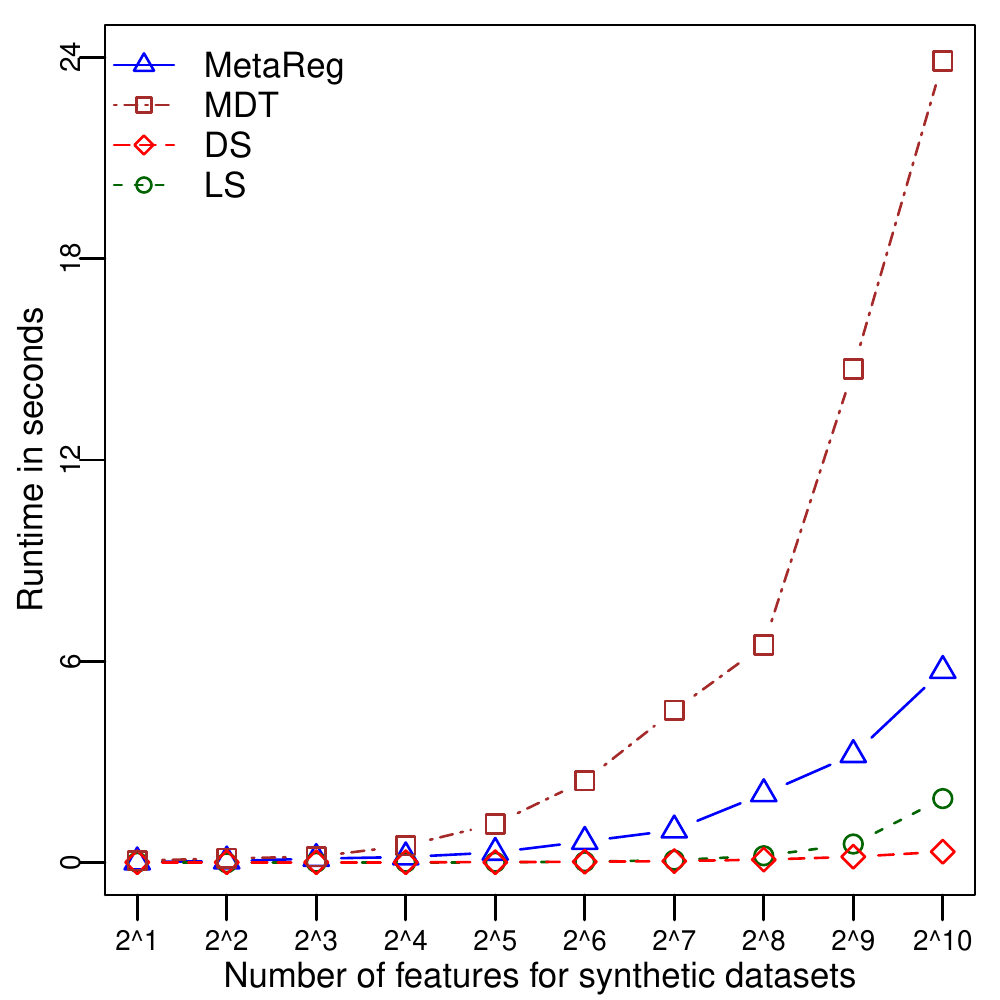}
	\caption{Empirical runtime scalability analysis of \texttt{MetaBags} resorting to samples (left panel) and features (right panel) size. Times in seconds.}
	\label{fig:scalability}
\end{figure*}
}
\afterpage{
\begin{table*}[!t]
\caption{Detailed predictive performance results, comparing base learners vs. \texttt{MetaBags} (top) and SoA methods in model integration vs. \texttt{MetaBags} - including variations (bottom). The results reported on the average and (std. error) of RMSE. The last rows depict the wins and losses based on the two-sample \textit{t}-test with the significance level.}
	\begin{center}\scalebox{0.82}{
		\begin{tabular}{l || r r r r || r}
        \hline
		Dataset & \texttt{SVR} & \texttt{PPR} & \texttt{RF} & \texttt{GB} &  \texttt{MetaBags}\\
		\hline
		R11  & 232.23($\scriptstyle5.2$) & 242.12($\scriptstyle8.1$)  & 229.18($\scriptstyle9.8$)  & 225.45($\scriptstyle6.5$)  &  \textbf{220.37($\scriptstyle4.6$)}
 \\
		R12  & 210.66($\scriptstyle3.9$) & 217.34($\scriptstyle5.4$) & 205.77($\scriptstyle2.5$) & 200.49($\scriptstyle6.7$) & \textbf{194.36($\scriptstyle4.2$)}  \\
		R21  & 225.81($\scriptstyle6.8$) & 240.12($\scriptstyle5.3$) & 235.55($\scriptstyle4.0$)  & \textbf{210.45($\scriptstyle7.7$)} &  218.87($\scriptstyle4.9$)\\
		R22  & 260.77($\scriptstyle5.9$)   & 269.19($\scriptstyle6.5$)  & 255.91($\scriptstyle5.3$)  & \textbf{248.11($\scriptstyle5.2$)} & 253.74($\scriptstyle2.9$)
  \\
        \hline
		C. Housing  & $9.6\mbox{e}^{4}(\scriptstyle 3.1\mbox{e}^{3}\textstyle)$ & $10.0\mbox{e}^{4}(\scriptstyle5.5\mbox{e}^{3}\textstyle)$ & $9.9\mbox{e}^{4}(\scriptstyle5.0\mbox{e}^{3}\textstyle)$ & $\mathbf{8.9\mbox{e}^{4}(\scriptstyle3.9\mbox{e}^{3}\textstyle)}$ & $9.2\mbox{e}^{4}(\scriptstyle3.8\mbox{e}^{3}\textstyle)$  \\
        Concrete & 12.19($\scriptstyle2.6$) & 15.81($\scriptstyle2.4$) & 17.14($\scriptstyle2.32$) & 14.54($\scriptstyle1.7$) & \textbf{11.73($\scriptstyle\mathbf{0.3}$)}  \\
		Delta A.  & $3.2\mbox{e}^{-4}(\scriptstyle1.2\mbox{e}^{-4}\textstyle)$  & $3.6\mbox{e}^{-4}(\scriptstyle2.4\mbox{e}^{-4}\textstyle)$ & $4.2\mbox{e}^{-4}(\scriptstyle2.3\mbox{e}^{-4}\textstyle)$ & $2.5\mbox{e}^{-4}(\scriptstyle1.8\mbox{e}^{-4}\textstyle)$ & $\mathbf{2.2\mbox{e}^{-4}(\scriptstyle9.9\mbox{e}^{-5}\textstyle)}$  \\
        2Dplanes  & 2.12($\scriptstyle0.1$)  & 2.57($\scriptstyle0.2$) & 2.36($\scriptstyle0.1$) & \textbf{2.01($\scriptstyle0.2$)} & \textbf{2.02($\scriptstyle0.1$)}  \\
        Elevators  & $6.2\mbox{e}^{-3}(\scriptstyle3.8\mbox{e}^{-4}\textstyle)$ & $6.3\mbox{e}^{-3}(\scriptstyle5.8\mbox{e}^{-4}\textstyle)$ & $6.4\mbox{e}^{-3}(\scriptstyle5.4\mbox{e}^{-4}\textstyle)$ & $6.7\mbox{e}^{-3}(\scriptstyle4.5\mbox{e}^{-4}\textstyle)$ &  $\mathbf{5.6\mbox{e}^{-3}(\scriptstyle5.8\mbox{e}^{-4}\textstyle)}$ \\
        Parkinsons  & $5.5\mbox{e}^{-2}(\scriptstyle7.0\mbox{e}^{-3}\textstyle)$ & $8.1\mbox{e}^{-2}(\scriptstyle7.0\mbox{e}^{-3}\textstyle)$ & $7.5\mbox{e}^{-2}(\scriptstyle6.0\mbox{e}^{-3}\textstyle)$ & $6.4\mbox{e}^{-2}(\scriptstyle6.0\mbox{e}^{-3}\textstyle)$ & $\mathbf{4.9\mbox{e}^{-2}(\scriptstyle5.0\mbox{e}^{-3}\textstyle)}$\\
  
		Physic.  & 3.78($\scriptstyle0.1$) & 3.90($\scriptstyle0.2$) & 4.99($\scriptstyle0.2$)
 & 3.70($\scriptstyle0.2$) & \textbf{3.64($\scriptstyle0.2$)}
 \\
        Pole  & 24.11($\scriptstyle7.2$)  & 30.24($\scriptstyle9.9$)  & 28.90($\scriptstyle5.9$)  & \textbf{18.54($\scriptstyle9.3$)}  & 20.12($\scriptstyle3.1$)
 \\
 		Puma32H  & $3.0\mbox{e}^{-2}(\scriptstyle3.0\mbox{e}^{-3}\textstyle)$ & $2.9\mbox{e}^{-2}(\scriptstyle3.0\mbox{e}^{-3}\textstyle)$ & $2.8\mbox{e}^{-2}(\scriptstyle3.0\mbox{e}^{-3}\textstyle)$ & $\mathbf{2.5\mbox{e}^{-2}(\scriptstyle6.0\mbox{e}^{-3}\textstyle)}$ & $2.7\mbox{e}^{-2}(\scriptstyle4.0\mbox{e}^{-3}\textstyle)$\\
		R. Wine   & \textbf{0.69($\scriptstyle0.0$)}  & 0.91($\scriptstyle0.0$)  & 0.88($\scriptstyle0.0$) & 0.79($\scriptstyle0.0$) & 0.70($\scriptstyle0.0$) \\
        W. Wine   & 0.70($\scriptstyle0.0$) & 0.86($\scriptstyle0.0$) & 0.76($\scriptstyle0.0$) & \textbf{0.62($\scriptstyle\mathbf{0.0}$)} &  \textbf{0.62($\scriptstyle0.0$)} \\
        CPU\_a.  & \textbf{5.18($\scriptstyle0.4$)} & 5.89($\scriptstyle0.3$) & 6.87($\scriptstyle0.2$) & 5.99($\scriptstyle0.3$) &  5.45($\scriptstyle0.2$) \\
		CPU\_s.  & 6.07($\scriptstyle0.1$) & 6.37($\scriptstyle0.3$) & 8.71($\scriptstyle0.4$)  & 6.31($\scriptstyle0.5$) & \textbf{5.12($\scriptstyle0.1$)} \\
		\hline \hline
		$\varnothing$ \textbf{Rank}  & 2.82 & 4.47 & 4.12 & 2.12 & \textbf{1.41}  \\
		\hline\hline\hline
        \textbf{Win/Loss}   & 10/0  &  11/0 &  4/0   &1/0 & N/A \\
        \hline
		\end{tabular}}	
    \scalebox{0.82}{
		\begin{tabular}{l || r r r | r  r || r}
        \hline
		Dataset &  \texttt{LS} & \texttt{DS} & \texttt{Best} & \texttt{MetaReg} & \texttt{MBwLM} & \texttt{MetaBags}\\
		\hline
		R11   & 230.25($\scriptstyle9.2$)  &222.37($\scriptstyle5.4$)  & 223.20($\scriptstyle8.9$)  & 240.67($\scriptstyle15.7$)  & 228.32($\scriptstyle7.0$) & \textbf{220.37($\scriptstyle4.6$)}  \\
		R12   & 215.32($\scriptstyle5.9$)& \textbf{190.80($\scriptstyle3.5$)} & 201.46($\scriptstyle4.3$) & 219.76($\scriptstyle19.2$) & 211.63($\scriptstyle7.9$) & 194.37($\scriptstyle4.2$)\\
		R21   & 234.37($\scriptstyle7.1$) & 221.98($\scriptstyle5.8$) & 226.27($\scriptstyle4.2$) & 249.31($\scriptstyle6.1$) & 220.01($\scriptstyle7.5$) & \textbf{218.87($\scriptstyle4.9$)} \\
		R22   & 260.71($\scriptstyle4.5$) & \textbf{250.37($\scriptstyle5.5$)} & 254.69($\scriptstyle5.2$) & 273.12($\scriptstyle5.4$) & 261.41($\scriptstyle4.3$) & 253.744($\scriptstyle2.9$) \\
        \hline
		C. Housing  & $9.8\mbox{e}^{4}(\scriptstyle4.3\mbox{e}^{3}\textstyle)$ &$9.3\mbox{e}^{4}(\scriptstyle3.2\mbox{e}^{3}\textstyle)$  & $9.5\mbox{e}^{4}(\scriptstyle4.6\mbox{e}^{3}\textstyle)$  & $9.8\mbox{e}^{4}(\scriptstyle4.1\mbox{e}^{3}\textstyle)$ & $9.5\mbox{e}^{4}(\scriptstyle3.7\mbox{e}^{3}\textstyle)$ & $\mathbf{9.2\mbox{e}^{4}(\scriptstyle3.7\mbox{e}^{3}\textstyle)}$ \\
        Concrete & 12.14($\scriptstyle0.3$)& \textbf{11.70($\scriptstyle\mathbf{0.3}$)}  & 12.00($\scriptstyle0.3$) & 12.57($\scriptstyle0.3$) & 11.91($\scriptstyle0.3$) & 11.73($\scriptstyle0.3$)	 \\
		Delta A.    & $2.8\mbox{e}^{-4}(\scriptstyle2.3\mbox{e}^{-4}\textstyle)$& $2.6\mbox{e}^{-4}(\scriptstyle1.8\mbox{e}^{-4}\textstyle)$ & $3.0\mbox{e}^{-4}(\scriptstyle1.8\mbox{e}^{-4}\textstyle)$ & $3.5\mbox{e}^{-4}(\scriptstyle1.7\mbox{e}^{-4}\textstyle)$ & $2.3\mbox{e}^{-4}(\scriptstyle1.8\mbox{e}^{-4}\textstyle)$ &$\mathbf{2.2\mbox{e}^{-4}(\scriptstyle9.9\mbox{e}^{-5}\textstyle)}$
 \\
        2Dplanes   & 2.35($\scriptstyle0.2$) & 2.26($\scriptstyle0.1$) & 2.34($\scriptstyle0.1$) & 2.39($\scriptstyle0.1$)  & 2.15($\scriptstyle0.1$) & \textbf{2.02($\scriptstyle0.1$)} \\
        Elevators  & $6.2\mbox{e}^{-3}(\scriptstyle5.5\mbox{e}^{-4}\textstyle)$ & $5.5\mbox{e}^{-3}(\scriptstyle7.0\mbox{e}^{-4}\textstyle)$ & $7.4\mbox{e}^{-3}(\scriptstyle6.1\mbox{e}^{-4}\textstyle)$ & $6.5\mbox{e}^{-3}(\scriptstyle4.6\mbox{e}^{-4}\textstyle)$ & $5.5\mbox{e}^{-3}(\scriptstyle6.8\mbox{e}^{-4}\textstyle)$& $\mathbf{5.7\mbox{e}^{-3}(\scriptstyle5.8\mbox{e}^{-4}\textstyle)}$  \\
        Parkinsons  & $5.9\mbox{e}^{-2}(\scriptstyle6.0\mbox{e}^{-3}\textstyle)$ & $6.5\mbox{e}^{-2}(\scriptstyle0.0\mbox{e}^{-3}\textstyle)$ & $6.2\mbox{e}^{-2}(\scriptstyle5.0\mbox{e}^{-3}\textstyle)$ & $5.4\mbox{e}^{-2}(\scriptstyle5.0\mbox{e}^{-3}\textstyle)$ & $5.1\mbox{e}^{-2}(\scriptstyle5.0\mbox{e}^{-3}\textstyle)$& $\mathbf{4.9\mbox{e}^{-2}(\scriptstyle5.0\mbox{e}^{-3}\textstyle)}$  \\
		Physic.   & \textbf{3.46($\scriptstyle0.2$)} & 3.74($\scriptstyle0.2$) & 3.79($\scriptstyle0.2$) & 3.80($\scriptstyle0.2$) & 3.71($\scriptstyle0.2$) & 3.64($\scriptstyle0.2$) \\
        Pole  & 26.22($\scriptstyle3.3$) & 22.15($\scriptstyle3.6$)  & 21.61($\scriptstyle3.3$) & 25.96($\scriptstyle4.3$) & 20.91($\scriptstyle3.6$) & \textbf{20.12($\scriptstyle\mathbf{3.1}$)} \\
        Puma32H  & $4.0\mbox{e}^{-2}(\scriptstyle5.0\mbox{e}^{-3}\textstyle)$ & $3.2\mbox{e}^{-2}(\scriptstyle6.0\mbox{e}^{-3}\textstyle)$ & $3.4\mbox{e}^{-2}(\scriptstyle4.0\mbox{e}^{-3}\textstyle)$ & $3.3\mbox{e}^{-2}(\scriptstyle6.0\mbox{e}^{-3}\textstyle)$ & $2.9\mbox{e}^{-2}(\scriptstyle4.0\mbox{e}^{-3}\textstyle)$& $\mathbf{2.7\mbox{e}^{-2}(\scriptstyle4.0\mbox{e}^{-3}\textstyle)}$  \\
		R. Wine   & 0.87($\scriptstyle0.0$) & 0.75($\scriptstyle0.0$) & \textbf{0.67($\scriptstyle0.0$)} & 0.76($\scriptstyle0.0$) & 0.70($\scriptstyle0.0$)
 & 0.70($\scriptstyle0.0$) \\
        W. Wine  & 0.82($\scriptstyle0.0$) & 0.63($\scriptstyle0.0$) & 0.67($\scriptstyle0.0$) & 0.76($\scriptstyle0.0$) & \textbf{0.61($\scriptstyle0.0$)} & 0.62($\scriptstyle0.0$) \\
        CPU\_a.   & 5.92($\scriptstyle0.2$) & 5.93($\scriptstyle5.5$) & 5.58($\scriptstyle0.3$) & 5.54($\scriptstyle0.3$) & 5.57($\scriptstyle0.2$) & \textbf{5.45($\scriptstyle\mathbf{0.2}$)} \\
		CPU\_s.  & 5.92($\scriptstyle0.2$) & 6.08($\scriptstyle0.3$) & 5.90($\scriptstyle0.2$) & 6.12($\scriptstyle0.3$) & 5.30($\scriptstyle0.2$)& \textbf{5.12($\scriptstyle0.2$)} \\
		\hline
        \hline
		$\varnothing$ \textbf{Rank} & 	4.76 & 3.06 & 3.88 & 5.12 & 2.65 & \textbf{1.47} \\
		\hline\hline\hline
        \textbf{Win/Loss}   & 4/0  & 2/0  &  4/0 &  6/0 & 0/0 & N/A \\
        \hline
		\end{tabular}	}
	\end{center}
    \label{results_base}
\end{table*}
}
	\section{Discussion}
The results, presented in Table \ref{results_base}, show that \texttt{MetaBags} outperforms existing SoA stacking methods. \texttt{MetaBags} is never statistically significantly worse than any of the other methods, which illustrates its generalization power. 

Fig. \ref{fig:summary_results} summarizes well the contribution of introducing bagging at the meta-level as well as the novel local landmarking meta-features, with average relative percentages of improvement in performance across all datasets of 12.73\% and 2.67\%, respectively. The closest base method is \texttt{GB}, with an average percentage of improvement of 5.44\%. However, if we weight this average by using the percentage of extreme target outliers of each dataset, the expected improvement goes up to 14.65\% - illustrating well the issues of \texttt{GB} depicted earlier in Fig. \ref{fig:intro_regression}-(b).

Fig. \ref{fig:scalability} also depicts the how competitive \texttt{MetaBags} can be in terms of scalability. Although not outperforming \texttt{DS} neither \texttt{LS}, we want to highlight that lazy learners have their cost in test time - while this study only covered the training stage.  
Based in the above discussion, (Q1-Q4) can be answered affirmatively.

One possible drawback of \texttt{MetaBags} may be its space complexity - since it requires to train/maintain multiple decision trees and models in memory. Another possible issue when dealing with low latency data mining applications is that the computation of some of the meta-features is not trivial, which may increase slightly its runtimes in test stage. Both issues were out of the scope of the proposed empirical evaluation and represent open research questions.

Like any other stacking approach, \texttt{MetaBags} requires training of the base models apriori. This pool of models need to have some diversity on their responses. Hereby, we explore the different characteristics of different learning algorithms to stimulate that diversity. However, this may not be sufficient. Formal approaches to strictly ensure diversity on model generation for ensemble learning in regression are scarce \cite{brown2005,mendes2012}. The best way to ensure such diversity within an advanced stacking framework like \texttt{MetaBags} is also an open research question.
	\section{Final Remarks}
This paper introduce \texttt{MetaBags}: a novel, practically useful stacking framework for regression. \texttt{MetaBags} uses meta-decision trees that perform on-demand selection of base learners at test time based on a series of innovative meta-features. These meta-decision trees are learned over data bootstrap samples, whereas the outputs of the selected models are combined by average. An exhaustive empirical evaluation, including 17 datasets and multiple comparison algorithms illustrates well the ability of \texttt{MetaBags} to address model integration problems in regression. As future work, we aim to study which factors may affect the performance of \texttt{MetaBags}, namely, at model generation level, as well as its time and spatial complexity in test time.

		
	\bibliographystyle{ACM-Reference-Format}
	\bibliography{BibTexLibrary}
		
	\end{document}